\documentclass[conference]{IEEEtran}
\IEEEoverridecommandlockouts
\usepackage{cite}
\usepackage{amsmath,amssymb,amsfonts}
\usepackage{algorithmic}
\usepackage{graphicx}
\usepackage{textcomp}
\usepackage{xcolor}
\usepackage{booktabs}
\usepackage{multirow}

\usepackage{hyperref}
\hypersetup{hidelinks}
\def\BibTeX{{\rm B\kern-.05em{\sc i\kern-.025em b}\kern-.08em
    T\kern-.1667em\lower.7ex\hbox{E}\kern-.125emX}}
\newenvironment{cyphercode}{%
\par\smallskip\noindent
\begin{minipage}{\columnwidth}
\hrule height 0.25pt
\vspace{1.5pt}
\scriptsize\ttfamily
\begin{tabular}{@{}p{0.98\columnwidth}@{}}
}{%
\end{tabular}
\vspace{1.5pt}
\hrule height 0.25pt
\end{minipage}
\par\smallskip
}
\begin{document}

\title{TCMaster: Confidence-Aware Querying and Workload-Guided\\Physical Design for Multi-Source Traditional Chinese Medicine Knowledge Graphs}

\author{
\IEEEauthorblockN{Zheng Chen\IEEEauthorrefmark{1}\IEEEauthorrefmark{4},
Yuzhu Li\IEEEauthorrefmark{2}\IEEEauthorrefmark{4},
Haoxuan Li\IEEEauthorrefmark{1}, Zhongde Zhang\IEEEauthorrefmark{3},
Lianshun Jin\IEEEauthorrefmark{3}, Peiwu Qin\IEEEauthorrefmark{3}\IEEEauthorrefmark{5}}
\IEEEauthorblockA{\IEEEauthorrefmark{1}Tsinghua University, China \quad
\IEEEauthorrefmark{2}Beijing University of Chinese Medicine, China\\
\IEEEauthorrefmark{3}Guangdong Provincial Laboratory of Traditional Chinese Medicine, China\\
\IEEEauthorrefmark{4}Equal contribution. \IEEEauthorrefmark{5}Corresponding author.\\
\{chen-z24, li-hx24\}@mails.tsinghua.edu.cn \quad 20220121029@bucm.edu.cn\\
\{zdzhang26, lsjin26\}@outlook.com, pwqin1979@gmail.com}
}

\maketitle

\begin{abstract}
Multi-source knowledge graphs (KGs) need query mechanisms that expose reliability and exploit domain structure. This paper presents TCMaster, a property-graph query substrate for confidence-aware traversal and workload-guided physical design over Traditional Chinese Medicine KGs. TCMaster integrates pharmacopoeias, prescriptions, molecular databases, and LLM-extracted micro-semantics into a KG with approximately 221K entities and 723K base edges. It annotates edges with provenance-level confidence, rewrites Cypher queries with confidence predicates, ranks multi-hop paths under PRODUCT, MIN, or weighted-average policies, and uses ontology skew through direction selection, herb-attribute bitmaps, and materialized shortcut edges. On Neo4j, direction selection improves attribute lookup by a factor of 1.47, shortcuts accelerate high-fanout target counting by a factor of 4.42, confidence filtering removes 39.3 percent of low-quality heterogeneous paths, and KG retrieval improves TCMbench QA accuracy by 20.0 percentage points.
\end{abstract}


\section{Introduction}

Traditional Chinese Medicine (TCM) contains heterogeneous knowledge across herbs, molecular ingredients, therapeutic targets, and classical prescriptions~\cite{b1,b2,b3,b4,b5}. Existing TCM knowledge bases are valuable repositories, but they are not query processing systems: they rarely expose edge reliability, provenance, or physical designs for ontology-skewed graph workloads.

Making a TCM knowledge graph (KG) queryable raises three data-management challenges. First, evidence reliability varies sharply across pharmacopoeias, curated prescription sources, molecular databases, LLM-extracted micro-semantics, and predicted links. Second, TCM ontologies are highly skewed: a few Nature, Flavor, Meridian, and Toxicity values connect to more than 15K Herb nodes, creating predictable traversal asymmetry. Third, clinical knowledge graph retrieval-augmented generation (KG-RAG), prescription safety inspection, and drug-target discovery require interactive multi-hop retrieval rather than static browsing. Existing resources such as TCMSP~\cite{b1}, TCMID~\cite{b2}/TCMID~2.0~\cite{b6}, SymMap~\cite{b3}, HERB~2.0~\cite{b4}, and ETCM~\cite{b5} do not jointly address these requirements. Table~\ref{tab:system-comparison} summarizes the gap.

\begin{table}[t]
\caption{Comparison with representative TCM resources. M, Rx, Syn, and Micro denote molecular, prescription, syndrome/symptom, and micro-semantic knowledge.}
\label{tab:system-comparison}
\centering
\scriptsize
\setlength{\tabcolsep}{2.0pt}
\begin{tabular}{lccccc}
\toprule
\textbf{System} & \textbf{Scope} & \textbf{Edge conf.} & \textbf{Path query} & \textbf{Opt.} & \textbf{Artifact}\\
\midrule
TCMSP & M & No & No & No & Web\\
TCMID & M/Disease & No & No & No & Web\\
SymMap & Syn & No & No & No & Web\\
HERB~2.0 & M & No & No & No & Web\\
ETCM & Rx & No & No & No & Web\\
OpenTCM & RAG/Diag. & No & No & No & Paper\\
\textbf{TCMaster} & M/Rx/Micro & Yes & Yes & Yes & Code+Data\\
\bottomrule
\end{tabular}
\end{table}

We present TCMaster, a property-graph substrate for \textit{confidence-aware traversal} and \textit{workload-guided physical design} over multi-source TCM KGs. TCMaster integrates 221K entities and 723K base edges, annotates every base edge with provenance-level confidence, rewrites Cypher queries with confidence predicates, ranks paths under PRODUCT, MIN, or weighted-average policies, and exploits ontology skew through direction selection, herb-attribute bitmaps, and materialized cross-layer shortcuts. It is not a new clinical LLM, diagnosis protocol, probabilistic database, or universal graph optimizer; it is a data-management substrate that makes reliability and ontology-induced asymmetry visible to query execution.

Experiments on Neo4j 4.4 show 5.2--12.0~ms latency across 1--4 hop workloads. Direction selection improves attribute lookup by 1.47$\times$, materialized shortcuts accelerate high-fanout target counting by 4.42$\times$, confidence filtering removes 39.3\% of low-quality heterogeneous paths at $\theta=0.18$, and downstream KG retrieval improves TCMbench QA accuracy by 20.0 percentage points (pp) over the LLM baseline. Cross-source validation reaches 93.7\% precision on herb-ingredient pairs and 95.5\% audited precision on LLM-extracted micro-semantics.

This paper makes the following contributions:
\begin{enumerate}
\item \textbf{A confidence-bounded path query processor.} We define PRODUCT, MIN, and weighted-average operational scoring policies, implement transparent Cypher rewriting for confidence constraints, and support incremental edge-confidence updates at 4.7--5.4~ms per edge without global recomputation.
\item \textbf{A workload characterization of ontology-skew physical design.} We exploit TCM-specific cardinality skew through traversal direction selection, compact attribute bitmaps, and materialized cross-layer shortcuts. The evaluation reports both gains and boundaries: 1.47$\times$ for attribute direction selection and 4.42$\times$ for high-fanout target counting, but limited benefit for bitmap-only and top-$k$ lookup workloads under Neo4j's native planner.
\item \textbf{A provenance-aware KG substrate and reproducible artifact.} We construct TCMaster-KG, integrating 221K entities and 723K base edges from four source types across a unified five-layer ontology, plus 28.75M materialized shortcut edges for reachability workloads. Every base edge carries a provenance-tagged confidence score, data validation reaches 93.7\% precision on herb-ingredient pairs and 95.5\% audited precision on LLM-extracted micro-semantics, and the submitted artifact provides code, query templates, result files, and a Neo4j dump snapshot for reproduction.
\end{enumerate}

\section{Preliminaries and Problem Definition}
\label{sec:prelim}

\subsection{Multi-Source TCM Knowledge Graph}

We model TCMaster-KG as a directed labeled multi-graph $G = (V, E, R, \mathit{src}, \mathit{dst}, \mathit{rel}, L)$, following standard graph database practices~\cite{b9,b11}. $V$ is the entity set, $E$ the edge identifier set, $R$ the relation type set, and $L$ assigns labels and properties. Each edge $e \in E$ has a source node $\mathit{src}(e)$, a destination node $\mathit{dst}(e)$, and a relation type $\mathit{rel}(e) \in R$. Node and edge properties store canonical names, identifiers, provenance fields, confidence values, and derived physical-design attributes. TCMaster-KG currently contains 221,225 entities and 722,671 base edges across 27 entity types and 23 relation types, organized in five knowledge layers (Table~\ref{tab:ontology}).

\begin{table}[htbp]
\caption{Five-layer ontology of TCMaster-KG. Edge counts summarize primary relation groups and are rounded.}
\begin{center}
\footnotesize
\setlength{\tabcolsep}{2pt}
\begin{tabular}{@{}p{0.20\columnwidth}p{0.50\columnwidth}p{0.20\columnwidth}@{}}
\toprule
\textbf{Layer} & \textbf{Content} & \textbf{Primary Edges}\\
\midrule
L1: Molecular & Herb--Ingredient--Target & 265K\\
L2: TCM Attributes & Nature, Flavor, Meridian, Toxicity & 45K\\
L3: Prescription & Prescription composition & 71K\\
L4: Micro-semantics & Processing, botany, efficacy, etc. & 236K\\
L5: Clinical & Prescription efficacy, indication & 20K\\
\bottomrule
\end{tabular}
\label{tab:ontology}
\end{center}
\end{table}

TCMaster models frequently queried TCM attributes as first-class graph nodes rather than repeated Herb properties. Thus attribute constraints become graph patterns, and Herb serves as the bridge across molecular, prescription, micro-semantic, and clinical layers. The rounded layer counts in Table~\ref{tab:ontology} cover the main relation groups; the full 722,671-edge graph also includes cross-layer mapping, normalization, and auxiliary relation types.

Source heterogeneity is preserved at the edge level. A relation imported from a molecular database and a relation extracted from text may share the same endpoint labels, but they remain distinguishable by source category, provenance pointer, and confidence score. This representation lets the query processor filter or rank evidence without duplicating the ontology for each data source.

\subsection{Confidence-Annotated Knowledge Graph}

A confidence-annotated KG extends $G$ with two functions: $\sigma: E \rightarrow [0,1]$ assigns a reliability score to each edge, and $\phi: E \rightarrow S$ traces each edge to its source. The source set $S$ contains the five levels listed in Table~\ref{tab:conf}; PREDICTED denotes outputs of knowledge graph embedding (KGE) models.

\begin{table*}[t]
\caption{Source-level confidence assignment}
\label{tab:conf}
\centering
\footnotesize
\setlength{\tabcolsep}{5pt}
\begin{tabular}{@{}p{0.27\textwidth}cp{0.58\textwidth}@{}}
\toprule
\textbf{Source Level} & \textbf{Conf.} & \textbf{Example Relations}\\
\midrule
AUTHORITATIVE & 0.95 & HAS\_INGREDIENT, TARGETS\\
AUTHORITATIVE\_PHARMA & 0.90 & HAS\_PROPERTY, HAS\_MERIDIAN\\
CURATED & 0.85 & HAS\_COMPONENT, HAS\_RX\_EFFICACY\\
LLM\_EXTRACTED & 0.70 & HAS\_BOTANY, PROCESSED\_BY\\
PREDICTED (KGE) & 0.30--0.60 & Link prediction outputs\\
\bottomrule
\end{tabular}
\end{table*}

AUTHORITATIVE edges come from pharmacopoeias and experimentally supported molecular databases; CURATED edges come from manually organized formulary sources; LLM\_EXTRACTED edges cover 11 relation categories with 95.5\% audited precision but receive a conservative 0.70 score to reflect hallucination risk~\cite{b8}; and PREDICTED edges are generated by KGE models~\cite{b16}. Unlike probabilistic databases or semiring provenance systems~\cite{b17,b18,b37}, TCMaster uses deterministic scores as execution annotations, exposing reliability ordering to threshold filters, path ranking, and provenance inspection.

\subsection{Confidence-Bounded Path Query}

Given a query pattern $Q$, an aggregation function $A: [0,1]^* \rightarrow [0,1]$, and a threshold $\theta \in [0,1]$, TCMaster returns all paths matching $Q$ whose aggregated confidence satisfies:
\begin{equation}
\begin{aligned}
\mathit{Result}(Q,A,\theta) =
\{p \in \mathit{Paths}_{G_c}(Q) \mid{}\\
A(\{\sigma(e): e \in p\}) \geq \theta\}.
\end{aligned}
\end{equation}
Three aggregation strategies are supported: PRODUCT ($\prod \sigma(e_i)$), penalizing long paths with any low-confidence edge; MIN ($\min \sigma(e_i)$), evaluating by the weakest link; and weighted average (WAVG) with $w_i = 1/(1 + \log i)$, giving higher weight to edges closer to the query start.

This definition separates candidate generation from result interpretation. Neo4j evaluates the graph pattern, TCMaster scores returned edge sequences, and $\theta$ denotes a query-time reliability threshold rather than a learned global cutoff. The system problem is to support these confidence-bounded path queries with predictable latency and auditable evidence: TCMaster must preserve provenance semantics, rewrite queries without changing logical answers except for the requested confidence or top-$k$ policy, and exploit stable ontology skew when it reduces traversal cost.

\section{System Architecture}
\label{sec:arch}

\begin{figure}[htbp]
\centerline{\includegraphics[width=\columnwidth]{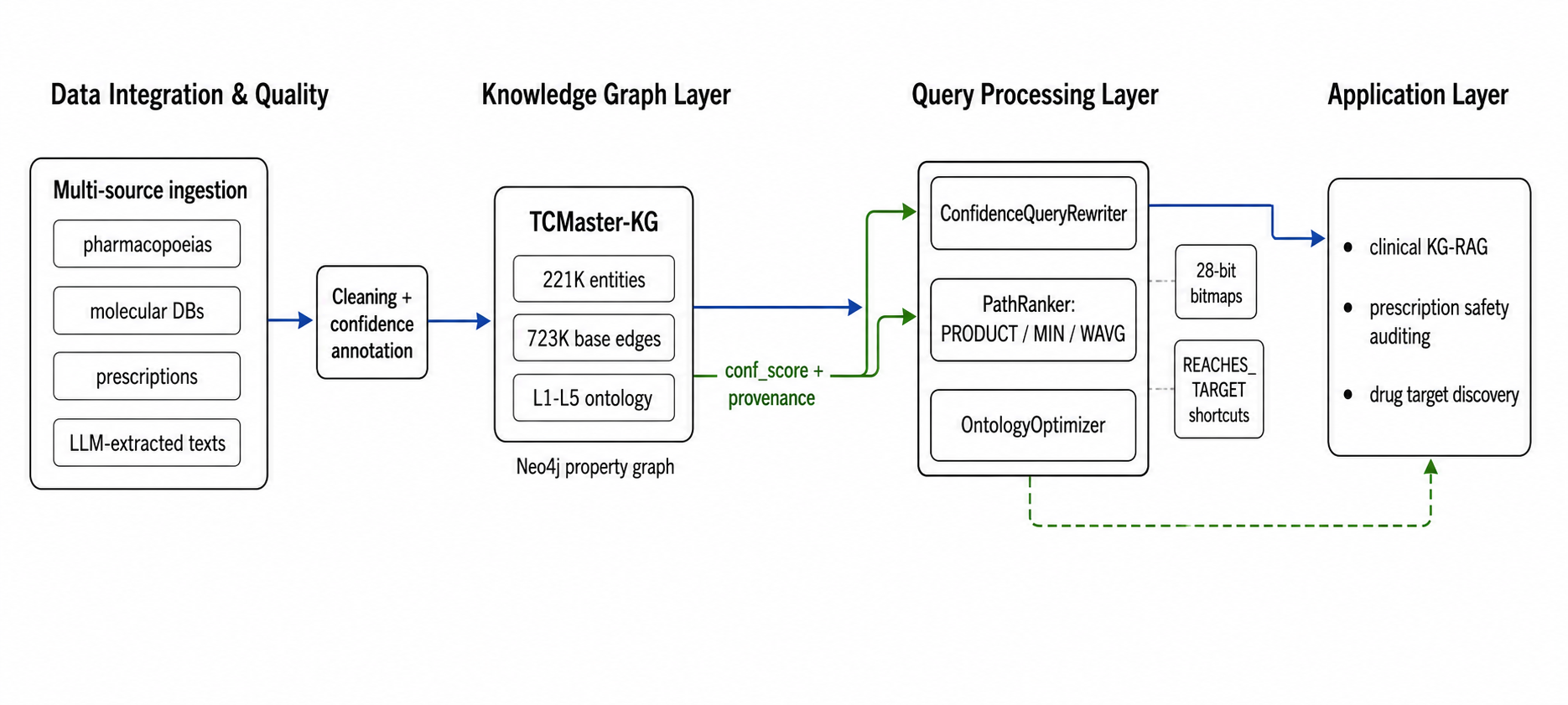}}
\caption{TCMaster system architecture from data ingestion and cleaning to confidence-annotated Neo4j storage, workload-guided query processing, and downstream application modes.}
\label{fig:arch}
\end{figure}

The formal model in \S2 defines what TCMaster-KG represents; Figure~\ref{fig:arch} shows how the system realizes it. TCMaster ingests and cleans multi-source data, stores it with confidence metadata and precomputed physical structures, processes confidence-aware queries with workload-guided rewrites, and serves application-facing modes. Two invariants guide the design: every base edge remains provenance complete, and physical structures are derived views that can be rebuilt or disabled without changing the logical ontology.

\subsection{Data Integration and Quality Layer}
\label{sec:data-integration}

TCMaster ingests four source categories: HERB~2.0 molecular relations~\cite{b4}, SymMap TCM-Western medicine links~\cite{b3}, classical prescription compositions (11,845 prescriptions), and LLM-extracted micro-semantics over 11 relation categories. External resources are treated as versioned snapshots with recorded dates, schema mappings, and checksums. Before ingestion, seven cleaning rules check identifier patterns, synonyms, mandatory keys, and numerical anomalies~\cite{b12,b13,b14}. Therapeutic category triples drop from 5,095 to 724 (14.2\% pass rate), while audited extraction precision rises from 91.2\% to 95.5\%. The layer outputs normalized edge tuples with canonical endpoints, relation type, source category, provenance pointer, validation status, and inherited confidence; discarded tuples remain in logs but do not enter the executable graph.

\subsection{Knowledge Graph Layer}
\label{sec:kg-layer}

All integrated data is stored in Neo4j 4.4 Community Edition using the property graph model. The unified ontology spans molecular interactions, TCM attributes, prescriptions, micro-semantics, and clinical prescription-level relations, with Herb entities connecting all five layers. TCMaster-KG augments the base schema with three derived structures: per-edge confidence and provenance tags (Section~\ref{sec:confidence}), per-herb 28-bit attribute bitmaps (Section~\ref{sec:bitmap}), and materialized REACHES\_TARGET shortcut edges for frequent Prescription$\rightarrow$Herb$\rightarrow$Ingredient$\rightarrow$Target traversals (Section~\ref{sec:shortcuts}). Confidence annotations define semantics, bitmaps summarize low-cardinality attributes, and shortcuts summarize high-fanout reachability.

These structures are kept separate because they answer different database questions. Confidence controls which evidence is admissible, bitmaps reduce local attribute filtering cost, and shortcuts reduce repeated cross-layer traversal cost. Disabling a bitmap or shortcut therefore changes performance experiments, not the logical ontology or provenance attached to base edges.

\subsection{Query Processing Layer}
\label{sec:query-layer}

The query processing layer provides confidence rewriting, workload-guided physical rewrites, and an optional natural-language-to-Cypher (NL2Cypher) template interface. At runtime, a Cypher template is bound to parameters, rewritten with edge-level predicates or path-level scoring code, optionally mapped to a reversed traversal, bitmap lookup, or shortcut relation, and executed by Neo4j. TCMaster returns ranked paths with the rewrite decision and provenance-bearing edge sequence. When a shortcut is used, the corresponding base-path pattern remains defined so that the result can be expanded or audited.

\subsection{Application Layer}
\label{sec:application-layer}

TCMaster supports clinical KG-RAG, prescription safety auditing, and drug-target discovery. All three modes consume the same result object: matched nodes, ordered edges, confidence values, source labels, and optional shortcut expansion. Applications therefore differ in templates and thresholds, not in hidden downstream schemas.

\section{Confidence-Aware Query Processing}
\label{sec:confidence}

\subsection{Confidence Annotation Model}

TCMaster-KG assigns a confidence score $\sigma(e) \in [0,1]$ and a source provenance label $\phi(e) \in S$ to every edge $e \in E$. The source set $S$ contains five levels: AUTHORITATIVE, AUTHORITATIVE\_PHARMA, CURATED, LLM\_EXTRACTED, and PREDICTED. The assignment follows a two-pass procedure implemented as Cypher transactions.

\textbf{Pass 1: Per-edge inheritance.} Some edges imported from upstream resources carry pre-existing confidence values. For instance, HERB~2.0 provides multi-source evidence weights for molecular interaction edges, typically in the range $[0.10, 0.45]$. Pass~1 preserves these values by setting \texttt{r.conf\_score = toFloat(r.confidence)} for all edges where \texttt{r.confidence IS NOT NULL}, and assigns the source level from the type-level mapping as a fallback via \texttt{COALESCE}:

\begin{cyphercode}
MATCH ()-[r:\{rtype\}]->()\\
WHERE r.confidence IS NOT NULL\\
SET r.conf\_score = toFloat(r.confidence),\\
\quad r.source\_level = COALESCE(r.source\_level, \$source)
\end{cyphercode}

\textbf{Pass 2: Type-level defaults.} Edges without per-edge confidence values are assigned the type-level constant from the assignment scheme in Table~\ref{tab:conf}:

\begin{cyphercode}
MATCH ()-[r:\{rtype\}]->()\\
WHERE r.conf\_score IS NULL\\
SET r.conf\_score = \$conf,\\
\quad r.source\_level = COALESCE(r.source\_level, \$source)
\end{cyphercode}

After both passes, per-type statistics are computed to verify annotation quality: \texttt{min(r.conf\_score)}, \texttt{max(r.conf\_score)}, \texttt{avg(r.conf\_score)} for each relationship type. The procedure processes all annotated relationship types sequentially, producing 722,671 annotated edges.

AUTHORITATIVE edges (0.95) derive from pharmacopoeias~\cite{b34} and experimentally supported molecular databases such as DrugBank~6.0 and TTD~\cite{b35,b36}. AUTHORITATIVE\_PHARMA edges (0.90) cover TCM pharmacopoeia attributes including Nature, Flavor, Meridian, and Toxicity. CURATED edges (0.85) come from manually organized formulary and prescription data sources. LLM\_EXTRACTED edges (0.70) cover the 11 micro-semantic categories; the automated cleaning pipeline (\S3.1) requires these edges to pass anchor-field validation before reaching the annotator. Although audit precision reaches 95.5\%, we conservatively assign 0.70 to reflect the non-trivial hallucination risk of LLM-generated content~\cite{b8}. PREDICTED edges (0.30--0.60) are reserved for KGE link prediction outputs, where model-specific confidence calibration determines the exact value.

The numeric values in Table~\ref{tab:conf} should be read as source-category reliability scores, not calibrated probabilities of clinical truth. They encode a stable ordering among evidence sources and make that ordering available to the execution layer. Per-edge inherited scores take precedence when upstream resources provide edge-level evidence weights; type-level constants are used only for relations without such scores. This avoids overwriting heterogeneous molecular evidence with a single type default while still ensuring that every edge has a query-visible reliability annotation. The threshold analysis in Section~\ref{sec:exp} evaluates how query results change across a range of $\theta$ values, rather than relying on a single calibrated cutoff.

\subsection{Confidence Semantics and Query Guarantees}

TCMaster treats confidence as an execution-visible edge annotation, not as a hidden data-cleaning score. For an edge $e$, $\sigma(e)$ summarizes source reliability and validation status, while $\phi(e)$ records the provenance category that explains why the score was assigned. A path confidence value is derived only after a query returns candidate paths; it is not stored as a global truth value for all possible paths. This distinction matters because edge confidence controls predicate pushdown, whereas path confidence controls ranking and post-filtering.

The thresholded query semantics are monotone with respect to $\theta$: if $\theta_1 \leq \theta_2$, every path returned under $\theta_2$ is also a candidate under $\theta_1$ before top-$k$ truncation. This property gives application developers a predictable reliability knob. Raising $\theta$ cannot introduce lower-confidence edges; it can only reduce or preserve the feasible path set. For PRODUCT and MIN aggregation, the path score is also monotone with respect to each edge score, so expert feedback that increases a validated edge confidence cannot lower the confidence of any path containing that edge.

TCMaster also preserves provenance continuity: each returned path includes edge sequence, confidence scores, and source labels, letting users distinguish pharmacopoeia-only paths from mixed LLM-extracted or predicted evidence. This is operational rather than statistical; PRODUCT is a conservative path score, not a possible-world probability, while MIN provides a weakest-link policy and WAVG preserves exploratory recall. This distinction is important for reviewer interpretation: TCMaster does not claim to solve probabilistic query evaluation, but it does expose reliability metadata at the same point where practitioners inspect graph evidence. The design therefore favors transparent execution behavior over hidden calibration assumptions.

\subsection{Confidence Query Rewriting}

Given a Cypher query $Q$ and threshold $\theta$, the \texttt{ConfidenceQueryRewriter} transparently injects confidence constraints. The rewriter first extracts all relationship variables from the query pattern using a regular expression that matches named Cypher relationships such as \texttt{[r]} and \texttt{[r:HAS\_INGREDIENT]}. For each extracted variable $r_i$, it generates the constraint $\texttt{r}_i\texttt{.conf\_score} \geq \theta$. All constraints are conjoined with AND and injected into the query.

If the original query contains a WHERE clause, constraints are prepended to it:
\begin{cyphercode}
MATCH (rx:Prescription)-[r1:HAS\_COMPONENT]->(h:Herb)\\
\quad -[r2:HAS\_INGREDIENT]->(i:Ingredient)\\
WHERE r1.conf\_score >= 0.8 AND r2.conf\_score >= 0.8\\
RETURN rx, h, i
\end{cyphercode}

If no WHERE clause exists, the rewriter locates the first occurrence of RETURN, WITH, ORDER, or LIMIT and inserts a WHERE clause before it:
\begin{cyphercode}
MATCH ...\\
WHERE r1.conf\_score >= 0.8\\
\quad AND r2.conf\_score >= 0.8\\
RETURN ...
\end{cyphercode}

This rewriting is transparent and deliberately narrow: it targets Cypher templates with explicit relationship variables in \texttt{MATCH} clauses, preserves existing predicates by conjunction, and leaves projection, grouping, ordering, and limits unchanged. For anonymous relationships, TCMaster first names the relationships. Each injected predicate maps to one relationship variable, keeping the transformation deterministic and executable by Neo4j. We intentionally avoid semantic rewrites that change return variables or aggregation clauses; this makes the transformation auditable and allows reviewers to compare original and rewritten queries in the artifact.

\subsection{Path Confidence Aggregation}

For multi-hop path queries, TCMaster extracts per-edge confidence scores and computes a single path-level confidence through the user-selected aggregation function. The \texttt{PathRanker} module appends a confidence extraction clause to the user's Cypher:

\begin{cyphercode}
\{user\_pattern\}\\
RETURN p, [r IN relationships(p) | r.conf\_score] AS scores\\
LIMIT 200
\end{cyphercode}

Three aggregation functions are supported. (1) \textbf{PRODUCT}: $\mathit{PathConf}(p) = \prod_{e \in p} \sigma(e)$. This penalizes long paths containing any low-confidence edge; a single LLM\_EXTRACTED edge at 0.70 reduces the aggregate to at most $0.95 \times 0.70 = 0.665$. PRODUCT is recommended for safety-critical scenarios. (2) \textbf{MIN}: $\mathit{PathConf}(p) = \min_{e \in p} \sigma(e)$. This evaluates the path by its weakest constituent edge, requiring every edge to individually satisfy the threshold. (3) \textbf{WAVG}: $\mathit{PathConf}(p) = \frac{\sum_{i=1}^{|p|} w_i \sigma(e_i)}{\sum_{i=1}^{|p|} w_i}$, where $w_i = 1/(1+\log i)$. This weighted average gives higher weight to edges closer to the query start, providing the highest recall and supporting exploratory analysis.

Paths below $\theta$ are discarded; the rest are ranked by descending $\mathit{PathConf}(p)$ and truncated to top-$k$ (default $k=20$). For $m$ paths of maximum length $\ell$, extraction and aggregation cost $O(m\ell)$ plus $O(m\log m)$ ranking. Edge-level predicates provide early pruning, while path aggregation is the final policy check; exploratory retrieval can set the edge threshold to zero and rank only after execution.

The design is intentionally conservative: instead of full probabilistic query semantics, TCMaster separates edge thresholding before execution from path ranking after execution. The same path can be inspected under PRODUCT, MIN, or WAVG without changing the graph or rebuilding indexes. This separation also supports different application modes: safety inspection can use strict edge filtering and PRODUCT ranking, while exploratory target discovery can retain more candidates and rank them post hoc.

\subsection{Incremental Confidence Updates}

TCMaster supports interactive confidence refinement through batch edge updates. When a domain expert validates or rejects a specific edge, the update sets both the new confidence score and marks the edge as \texttt{source\_level = USER\_VERIFIED}:

\begin{cyphercode}
MATCH (a)-[r:\{rel\_type\}]->(b)\\
WHERE id(a) = \$sid AND id(b) = \$eid\\
SET r.conf\_score = \$conf, r.source\_level = 'USER\_VERIFIED'
\end{cyphercode}

Updates are executed as Neo4j write transactions, so a feedback batch either commits all confidence and provenance changes or leaves the previous annotation state intact. Since confidence is stored on edges, update cost depends on the number of matched relationships rather than graph size. Batch updates of 100 or more edges achieve stable per-edge latency of 4.7--5.4 ms, with 1,000-edge batches completing within 5 seconds. The update does not trigger global recomputation; only the directly modified edges are affected. For materialized shortcuts that depend on updated edges, a staleness flag is set via the \texttt{check\_shortcut\_staleness} method, and affected shortcuts are lazily recomputed on the next materialization cycle.

\section{Workload-Guided Physical Design for Ontology-Skewed Graph Queries}
\label{sec:optimization}

Clinical workloads in TCM are dominated by three query patterns: (i) \textit{attribute lookup}---given a herb, retrieve its Nature, Flavor, Meridian, and Toxicity values; (ii) \textit{attribute-constrained search}---find herbs matching specific attribute combinations (e.g., Cold and Bitter and targeting Liver); and (iii) \textit{cross-layer traversal}---follow multi-hop paths such as Prescription$\rightarrow$Herb$\rightarrow$Ingredient$\rightarrow$Target for drug discovery or Prescription$\rightarrow$Herb$\rightarrow$Warning for safety auditing. These patterns share a structural bottleneck: each involves traversing from high-cardinality entity nodes (15K+ herbs) to low-cardinality attribute nodes (5--12 values) or across three knowledge layers. General-purpose graph databases use cost-based optimizers that treat all nodes as having comparable cardinalities, missing the optimization opportunities created by TCM's extreme ontology skew. We exploit this skew through three techniques: cardinality-based direction selection (Section~\ref{sec:direction}), bitmap-based attribute pruning (Section~\ref{sec:bitmap}), and materialized cross-layer shortcuts (Section~\ref{sec:shortcuts}).

\subsection{Cardinality Skew in TCM Ontology}
\label{sec:skew}

TCM ontologies exhibit orders-of-magnitude cardinality skew between entity types. Attribute nodes have very small, fixed cardinalities: TCMMeridian (12 nodes), TCMProperty (5 nodes for Cold, Hot, Warm, Cool, Neutral), TCMFlavor (7 nodes for Sour, Bitter, Sweet, Pungent, Salty, Bland, Astringent), TCMToxicity (4 nodes for Non-toxic, Slightly-toxic, Toxic, Highly-toxic). In contrast, Herb nodes number 15,092, Ingredient nodes 44,595, and Target nodes 15,515. This cardinality skew---attribute nodes being three orders of magnitude smaller than entity nodes---creates three optimization opportunities that general graph engines do not exploit by default.

\subsection{Cardinality-Based Direction Selection}
\label{sec:direction}

When a Cypher query constrains TCM attributes (e.g., ``find herbs that are Cold in Nature and target the Liver meridian''), the default traversal from the Herb side starts from 15K candidate nodes and fans out to attribute nodes. Our optimizer detects such Herb-to-attribute patterns and reverses the traversal direction:

\begin{cyphercode}
Original query:\\
MATCH (h:Herb)-[:HAS\_MERIDIAN]->\\
\quad (m:TCMMeridian \{name:'GanJing'\})\\[1pt]
Rewritten query:\\
MATCH (m:TCMMeridian \{name:'GanJing'\})\\
\quad <-[:HAS\_MERIDIAN]-(h:Herb)
\end{cyphercode}

This transformation reduces the initial search space from 15K+ Herb nodes to a single Meridian node. The optimizer applies this rewrite statically before query execution; the Cypher query planner then uses the low-cardinality starting point to bound the search. For queries involving multiple attribute constraints, all direction reversals are applied simultaneously. In our experiments, this optimization is transparent---users write queries in the natural Herb$\rightarrow$Attribute direction, and the system automatically selects the efficient Attribute$\rightarrow$Herb traversal.

\subsection{Bitmap-Based Attribute Pruning}
\label{sec:bitmap}

The TCM attribute space is small and closed in the current ontology snapshot: 5 properties, 7 flavors, 12 meridians, and 4 toxicity levels, yielding 28 distinct values in total. We exploit this by precomputing a 28-bit attribute bitmap for every Herb node, encoding the presence of each attribute value as a single bit.

\textbf{Bit layout.} The 28-bit integer uses four contiguous blocks: Nature (bits 0--4), Flavor (5--11), Meridian (12--23), and Toxicity (24--27). The layout follows the ontology rather than observed frequency, which keeps the encoding stable across graph snapshots and makes each bit interpretable during debugging. If later releases add controlled values, the representation can allocate additional bits or a second integer word without changing query semantics.

\textbf{Encoding.} For each Herb node $h$, $\mathcal{B}(h) = \bigvee_{\mathit{attr} \in \mathit{Attrs}(h)} (1 \ll \mathit{index}(\mathit{attr}))$. The bitmap is computed once during ingestion, permits multiple values per dimension, and is stored directly on Herb nodes. In the current snapshot, 11,014 Herb nodes receive at least one bitmap property, and the builder performs 31,426 herb-dimension bitmap updates across the four attribute dimensions. The non-mutual-exclusive design matters for TCM: a herb may have several flavors or meridians, so a categorical one-hot representation would either lose information or require repeated property values. The bitmap keeps multi-label attributes compact while leaving the original graph edges intact for provenance and explanation.

\textbf{Query-time pruning.} A query requiring a 2-hop traversal (Herb$\rightarrow$Meridian$\rightarrow$filter by name) is reduced to a 1-hop lookup plus a constant-time bitwise AND:
\begin{center}
\texttt{WHERE h.meridian\_bitmap \& (1 << k) > 0}
\end{center}
where $k$ is the target meridian index. For multi-attribute queries, several such constant-time checks replace repeated 2-hop traversals. This does not make bitmap filtering universally faster than graph traversal; rather, it creates a cheap local test once a candidate herb set has already been produced. Section~\ref{sec:exp} therefore reports both useful and limited cases instead of presenting bitmaps as a general-purpose index.

\subsection{Materialized Cross-Layer Shortcuts}
\label{sec:shortcuts}

The most frequent clinical query pattern traverses three knowledge layers: \texttt{Prescription $\rightarrow$ Herb $\rightarrow$ Ingredient $\rightarrow$ Target}. This 3-hop pattern is evaluated repeatedly in clinical QA and drug discovery workloads. TCMaster materializes this pattern as a direct shortcut edge \texttt{Prescription $\xrightarrow{\text{REACHES\_TARGET}}$ Target}.

\textbf{Construction.} Shortcuts are created by batched Cypher processing over 200 prescriptions per batch: expand the 3-hop path, deduplicate at the (Prescription, Target) level, skip existing shortcuts, and create \texttt{REACHES\_TARGET} with metadata. Each shortcut stores \texttt{conf\_score}=0.8075, the PRODUCT of CURATED and two AUTHORITATIVE edges, and \texttt{source\_level=MATERIALIZED}. The edge is not a new biomedical assertion; it is a physical design artifact summarizing reachability through existing evidence paths. We keep the source level explicit so applications can distinguish materialized reachability from base pharmacological facts.

\textbf{Maintenance.} Shortcuts are timestamped, lazily invalidated after constituent-edge updates, and recomputed in the next materialization cycle. This matches snapshot-oriented TCM sources; high-update deployments should rebuild shortcuts offline or disable them. Creating 28.75M shortcut edges takes about 74 minutes for the complete graph; the longer target-reachability stage also includes validation, indexing, and export checks. This cost is acceptable for offline release preparation but not for per-query construction, which is why TCMaster treats shortcuts as a workload-guided physical structure rather than a dynamic query rewrite alone.

\subsection{Rewrite Applicability and Cost Intuition}

The optimizer is rule-based, but the rules are not applied blindly. Each rewrite is tied to a structural precondition and a cost intuition. Direction reversal is applicable when a query binds one or more low-cardinality ontology values and asks for matching high-cardinality entities. Its benefit comes from reducing the starting frontier before expansion. It is therefore useful for Herb-Attribute and Prescription-Herb-Attribute workloads, but it provides little benefit when the query already starts from a selective entity identifier.

Bitmap pruning applies to the closed TCM attribute vocabulary, replacing repeated edge expansion with integer predicates. It helps multi-attribute filtering and repeated screening, but is less useful for single top-$k$ lookups where Neo4j already exploits selective node access. This boundary is important: a property predicate can reduce arithmetic cost while also changing the planner's available access path. TCMaster therefore enables bitmap predicates only when the query shape suggests repeated local attribute checks, not when a selective ontology node is already available as the starting point.

Shortcut materialization applies when a long traversal is reused as a reachability primitive. TCMaster materializes Prescription$\rightarrow$Target paths for drug discovery, clinical QA, and target-count aggregation; the benefit is largest for high-fanout counting and grouping. The cost is storage and maintenance for 28.75M shortcut edges, so infrequent exploratory traversals remain on the native graph engine. In this sense, the optimizer is closer to a conservative physical-design policy than a universal graph query optimizer: it exposes a small number of domain-stable structures and uses them only when the workload makes the trade-off favorable.

\section{Implementation}

TCMaster is implemented in Python 3.9 with approximately 12K lines of code across data pipeline, confidence annotation, query processing, and optimization modules. TCMaster-KG runs on Neo4j 4.4 Community Edition and stores three added structures: per-edge confidence/provenance properties, 28-bit Herb attribute bitmaps, and 28.75M materialized REACHES\_TARGET shortcuts. Neo4j is configured with 8~GB heap and 16~GB page cache on NVMe storage.

\textbf{System organization.} The implementation follows four modules that match the logical architecture in Fig.~\ref{fig:arch}. The Extract-Transform-Load (ETL) module normalizes heterogeneous source files into typed entity and relation tables. The confidence module annotates each edge, writes provenance properties, and exports per-relation statistics for audit. The query module rewrites Cypher templates and ranks paths. The optimization module builds bitmap and shortcut structures offline, then applies rewrite rules only when a query matches a supported pattern.

\textbf{Query pipeline.} TCMaster processes each query in three stages. The \texttt{ConfidenceQueryRewriter} names relationship variables and injects \texttt{r.conf\_score >= $\theta$} predicates. The \texttt{OntologyOptimizer} applies direction reversal, bitmap filtering, and shortcut substitution when the query pattern matches the corresponding workload. Neo4j executes the rewritten query, and the \texttt{PathRanker} computes aggregate path confidence, filters by $\theta$, and returns top-$k$ paths.

\textbf{Extraction and validation.} LLM extraction uses DeepSeek-V3 through SiliconFlow with batched processing, retries, and deterministic MD5-based subsampling (seed~42). The audit model checks triples against source text using temperature~0.1 and JSON output. KGE validation uses PyKEEN~1.10/PyTorch~2.0, six embedding models over S1--S4, 3 seeds, dimension~200, and filtered rank-based evaluation; Relational Graph Convolutional Network (R-GCN) is an auxiliary baseline. We report KGE as structural validation rather than as a deployed inference module, because link prediction scores are sensitive to relation heterogeneity and negative sampling choices.

\textbf{Benchmarking and ETL.} Query latency is measured on Intel Xeon Gold 6342, 256~GB RAM, and NVMe SSD with 100 sampled herbs/prescriptions, 3 warmup iterations, and 10 measured iterations per sample. The full ETL pipeline takes about 36~hours: 8~h ingestion, 2~h cleaning, 10~h confidence annotation, 12~h target-reachability preprocessing, and 4~h bitmap construction. These construction costs are separated from online latency measurements. The artifact contains scripts, templates, configurations, results, and reproduction instructions so reviewers can inspect short-running verification scripts without rebuilding the full KG.

\textbf{Reproducibility controls.} All experiments use fixed random seeds, pinned query templates, and materialized result files. Expensive construction steps, including shortcut generation and KGE training, are separated from short verification scripts so reviewers can inspect reported tables without rerunning the full pipeline. For large KG snapshots, the artifact records schema definitions, loader commands, checksum files, and sampled data slices, while the public release will provide downloadable snapshots for full reconstruction.

\section{Experimental Evaluation}
\label{sec:exp}

We organize the evaluation around five research questions, each tied to a database-system claim. RQ1 evaluates graph-query latency under workload-guided rewrites: when do direction selection, bitmap predicates, and shortcut substitution reduce execution time relative to Neo4j's native planner? RQ2 evaluates execution behavior under sampled KG scale-up on the core workload layers. RQ3 evaluates whether extraction, cleaning, and confidence annotation produce auditable graph facts with explicit provenance. RQ4 treats link prediction as an auxiliary KG structural diagnostic rather than as the main system contribution. RQ5 evaluates downstream retrieval as an application-level validation of the graph substrate, not as evidence for a new clinical model.

\subsection{Experimental Setup}

TCMaster-KG contains 221,225 entities, 722,671 base edges, and 28.75M materialized shortcut edges. KGE ablations use S1--S4 layer settings, six embedding baselines (TransE, RotatE, ComplEx, DistMult, TuckER, MuRE), and auxiliary R-GCN where available, with dimension 200 and seeds 42, 123, and 456. Query benchmarks run on Neo4j 4.4 with 8~GB heap. Each workload samples 100 start entities and repeats three runs after warmup. Unless stated otherwise, \textit{Base} is Neo4j 4.4's native planner on the same graph, constraints, and indexes, with TCMaster rewrites, bitmaps, and shortcut substitution disabled.

\subsection{Workload Characterization}

The benchmark suite is organized by operator stress rather than hop count alone: attribute-cardinality skew, cross-layer traversal, grouping, high-fanout matching, and safety audit. Table~\ref{tab:workload} summarizes the workloads.

This workload design is intentionally mixed. Simple 1-hop and 2-hop queries test whether confidence predicates and ontology rewrites add overhead to already selective traversals. Cross-layer 3-hop and 4-hop queries test the setting where provenance and path confidence matter most, because each answer may combine prescription, ingredient, target, disease, and predicted evidence. Aggregation and high-fanout workloads test whether the physical structures are useful beyond top-$k$ lookup. The combination prevents the evaluation from reporting only favorable cases: the same benchmark suite includes workloads where TCMaster should help and workloads where Neo4j's native planner is already competitive.

\begin{table}[t]
\caption{Characterization of the query workload. Rx and Ingr. denote prescription and ingredient.}
\label{tab:workload}
\centering
\footnotesize
\setlength{\tabcolsep}{2.2pt}
\begin{tabular}{lllr}
\toprule
\textbf{Query} & \textbf{Pattern} & \textbf{Operator stress} & \textbf{Avg. out}\\
\midrule
Q1 & Herb-Attr & 1-hop lookup & 3.03\\
Q2 & Rx-Herb-Attr & 2-hop join & 5.27\\
Q3 & Rx-Herb-Ingr.-Target & shortcut path & 91.10\\
Q4 & Rx-Herb-Ingr.-Disease & long traversal & 91.02\\
Q5 & Rx-Herb-Meridian & aggregation & 6.28\\
Q6 & Rx-Herb-Rx & pattern match & 22.00\\
Q7 & Rx-Herb-Safety & safety audit & 5.27\\
\bottomrule
\end{tabular}
\end{table}

\subsection{Query Performance}

Table~\ref{tab:query} summarizes latency for seven representative clinical workloads.

\begin{table}[htbp]
\caption{Query latency for representative clinical workloads}
\begin{center}
\begin{tabular}{lccr}
\toprule
\textbf{Query} & \textbf{Hops} & \textbf{Mean (ms)} & \textbf{P95 (ms)}\\
\midrule
Q1: Herb attribute lookup & 1 & 8.41 & 9.64\\
Q2: Prescription property & 2 & 5.23 & 6.49\\
Q3: Prescription$\rightarrow$Target & 3 & 6.70 & 8.89\\
Q4: Disease association & 4 & 7.12 & 9.77\\
Q5: Aggregation & 2 & 5.27 & 6.54\\
Q6: Pattern matching & var. & 12.04 & 17.52\\
Q7: Safety audit & 2 & 5.40 & 6.56\\
\bottomrule
\end{tabular}
\label{tab:query}
\end{center}
\end{table}

Key findings: (1) Latency is not monotonic with hop count: the 4-hop Q4 (7.12 ms) is faster than the 1-hop Q1 (8.41 ms), because latency is dominated by index lookup and intermediate cardinality rather than path length alone. (2) Cold/warm cache comparison for Q3 shows negligible difference (6.74 ms vs.\ 6.73 ms), indicating the graph is effectively memory-resident. (3) Q6 (pattern matching with high fanout) shows the highest latency, identifying this as the remaining bottleneck.

The P95 values refine the same conclusion. Six of seven workloads remain below 10~ms at P95, while Q6 reaches 17.52~ms because variable-length pattern matching must explore a broader candidate frontier before returning bounded results. This spread matters for interpreting TCMaster as a query substrate: the system is interactive for the tested clinical lookup, safety, and target-discovery templates, but the latency profile is still workload dependent. The results also show why we report representative templates rather than only aggregate averages. Q2, Q5, and Q7 have similar mean latency despite different semantics because their traversals are anchored by selective prescription or safety nodes; Q1 is slower because attribute lookup can start from higher-degree Herb neighborhoods unless direction selection is applied. Thus the query-performance result supports a bounded claim: confidence-aware graph retrieval can be served at interactive latency for the evaluated workloads, while high-fanout pattern matching remains the main optimization target.

\subsection{Optimization Evaluation}

We evaluate the three workload-guided physical-design choices on two classes of graph-query workloads. The first class mirrors interactive top-$k$ lookup, where Neo4j can often terminate early after finding 100 results. The second class targets high-fanout complete expansion and aggregation, where the execution layer must avoid repeatedly materializing large intermediate paths. Table~\ref{tab:ablation} reports the measured results.

\begin{table}[t]
\caption{Workload-guided physical-design results}
\label{tab:ablation}
\centering
\footnotesize
\setlength{\tabcolsep}{2.5pt}
\begin{tabular}{lrrrr}
\toprule
\textbf{Workload} & \textbf{Base} & \textbf{Opt.} & \textbf{Speedup} & \textbf{Avg. out}\\
\midrule
Direction selection & 9.40 & 6.41 & 1.47$\times$ & 100\\
Shortcut top-$k$ lookup & 13.62 & 11.64 & 1.17$\times$ & 100\\
Bitmap single attribute & 15.99 & 15.53 & 1.03$\times$ & 200\\
Combined top-$k$ lookup & 10.73 & 16.45 & 0.65$\times$ & 100\\
High-fanout target count & 51.19 & 11.57 & 4.42$\times$ & 6555\\
High-fanout target enum. & 336.08 & 283.43 & 1.19$\times$ & 6555\\
\bottomrule
\end{tabular}
\vspace{2pt}

\footnotesize Latency is mean ms. Base is Neo4j native planning on the same graph and indexes, with the corresponding TCMaster transformation disabled. High-fanout workloads use 20 prescriptions with 5K--8K reachable targets and 5 repeated runs.
\end{table}

\textbf{Direction selection.} Reversing traversal from Herb$\rightarrow$Attribute to Attribute$\rightarrow$Herb improves property lookup from 9.40~ms to 6.41~ms (1.47$\times$), confirming that low-cardinality ontology nodes can bound the search before expansion to Herb nodes.

\textbf{Materialized shortcuts.} For top-$k$ target lookup, shortcuts improve latency only modestly (13.62~ms to 11.64~ms, 1.17$\times$) because both plans stop after 100 targets. The benefit is larger for complete high-fanout expansion: target counting over prescriptions with 5K--8K reachable targets improves from 51.19~ms to 11.57~ms (4.42$\times$), while complete enumeration improves from 336.08~ms to 283.43~ms (1.19$\times$).

\textbf{Bitmap pruning.} Single-attribute bitmap filtering provides little benefit over Neo4j's native traversal (1.03$\times$), and multi-attribute probes are mixed because scanning Herb nodes with bitmap predicates can be more expensive than starting from selective attribute nodes. We therefore treat bitmaps as a compact representation for repeated local checks rather than the main accelerator.

\textbf{Combined optimization.} Shortcut lookup plus a confidence predicate is slower than unoptimized top-$k$ traversal (0.65$\times$) because \texttt{REACHES\_TARGET} edges already carry high confidence, so the predicate adds cost without pruning many edges. TCMaster therefore applies rewrites selectively instead of assuming that transformations compose beneficially.

Overall, the optimizer results support a narrower but defensible claim: TCM ontology skew can be exploited effectively when the workload exposes the corresponding bottleneck. Direction selection benefits low-cardinality attribute lookup, and materialized shortcuts substantially improve high-fanout complete expansion. For top-$k$ lookup workloads already optimized by Neo4j, the gains are limited.

\textbf{Selective rewrite policy.} TCMaster applies transformations only when their workload assumptions hold: direction reversal for Herb--attribute patterns with ontology cardinality below 32, shortcut substitution for aggregate counts or complete enumeration over Prescription$\rightarrow$Herb$\rightarrow$Ingredient$\rightarrow$Target paths, bitmap predicates for repeated local herb-attribute checks, and no shortcut-plus-confidence composition for already high-confidence shortcuts unless pruning is expected. This is not a full cost model, but an auditable policy derived from the ablation; the artifact includes Base/Opt. query templates.

Two design implications follow from the ablation. First, materialization should be justified by output cardinality, not merely by path length. The shortcut path is only modestly faster for top-$k$ lookup because both plans can stop early, but it is much faster when the query must count all reachable targets. Second, reliability predicates are not free: when a shortcut relation already has nearly uniform high confidence, adding a threshold predicate can reduce planner flexibility without removing many edges. These observations motivate keeping the rewrite policy explicit and auditable rather than hiding it behind an opaque cost heuristic.

\subsection{Core-Layer Scalability}

We evaluate query latency at four scales of the core L1--L3 subgraph (25\%, 50\%, 75\%, 100\%) by sampling herb subsets and re-importing subgraphs into a clean Neo4j instance. Results are shown in Table~\ref{tab:scale}.

\begin{table}[htbp]
\caption{Query latency (ms) vs.\ core-layer graph scale}
\begin{center}
\begin{tabular}{lrrrrr}
\toprule
\textbf{Scale} & \textbf{Nodes} & \textbf{Edges} & \textbf{Q1} & \textbf{Q3} & \textbf{Q5}\\
\midrule
25\% & 26,575 & 55,263 & 7.98 & 6.50 & 11.73\\
50\% & 41,755 & 113,000 & 11.67 & 6.43 & 10.84\\
75\% & 53,698 & 165,745 & 8.08 & 2.83 & 5.59\\
100\% & 64,254 & 223,696 & 6.32 & 1.99 & 5.80\\
\bottomrule
\end{tabular}
\label{tab:scale}
\end{center}
\end{table}

In the controlled L1--L3 scale benchmark, latency remains bounded but not monotonic. Q3 decreases from 6.50 ms at 25\% to 1.99 ms at 100\%, consistent with favorable index utilization in the sampled subgraphs. The temporary Q1 increase at 50\% (11.67 ms vs.\ 7.98 ms at 25\%) reflects over-sampling herbs with dense attribute out-degree. This is a stability check for latency-critical paths, not a claim about arbitrary graph scaling; extending controlled scale-up to L4--L5 remains future work.

The non-monotonicity is worth reporting because it prevents a misleading scalability narrative. In Neo4j, sampled subgraphs can change both data volume and selectivity; a larger sample may expose more index-friendly anchors or more selective attribute values. We therefore use this experiment as a bounded robustness check for the core query layer, while the full 221K-node graph remains the primary evaluation setting for reported workload latency.

\begin{figure}[htbp]
\centering
\includegraphics[width=\columnwidth]{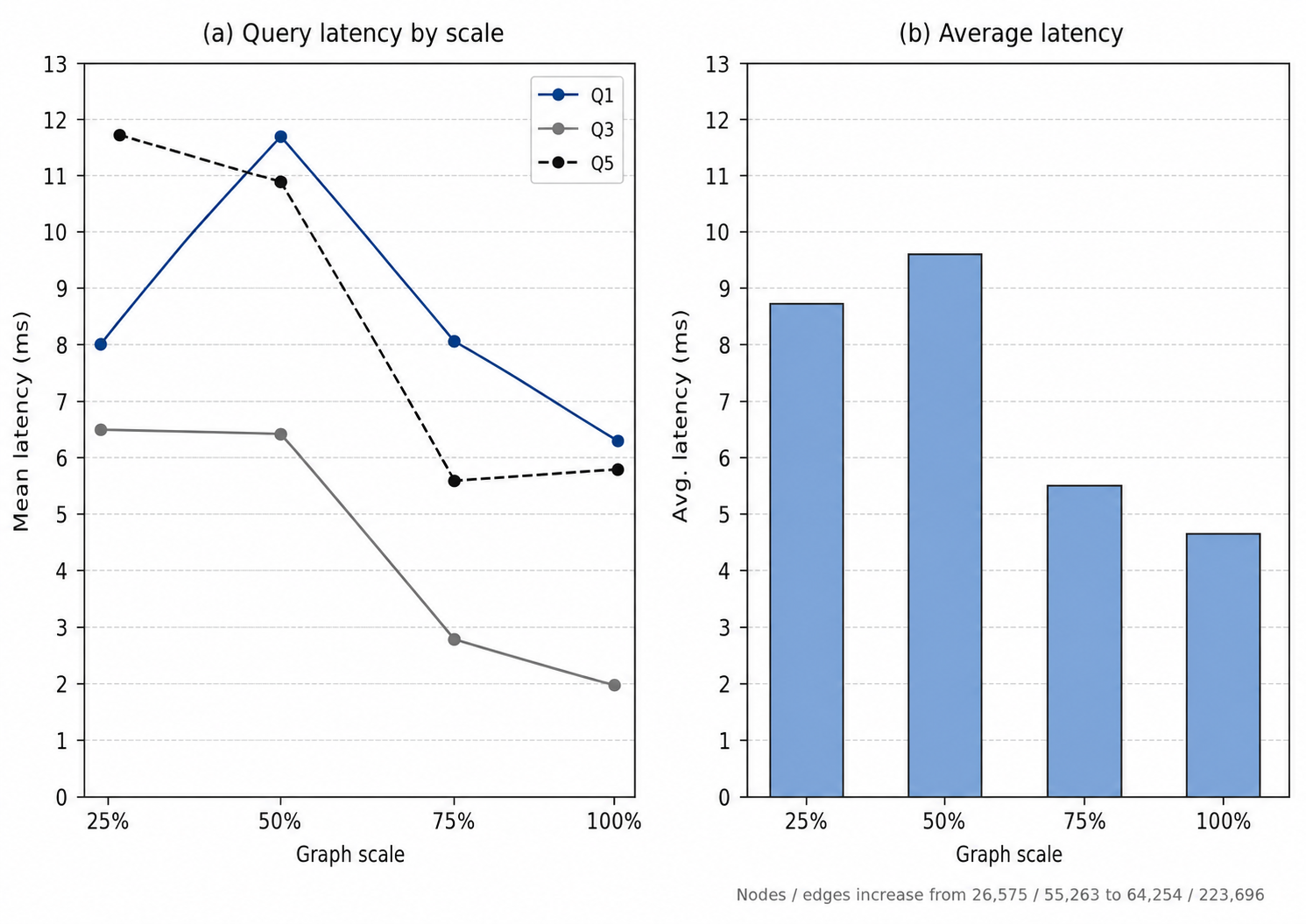}
\caption{Core-layer scalability over 25--100\% sampled L1--L3 graph scales, showing bounded latency variation for Q1, Q3, and Q5.}
\label{fig:scalability}
\end{figure}

\subsection{Data Quality and Extraction}

\begin{figure}[htbp]
\centering
\includegraphics[width=\columnwidth]{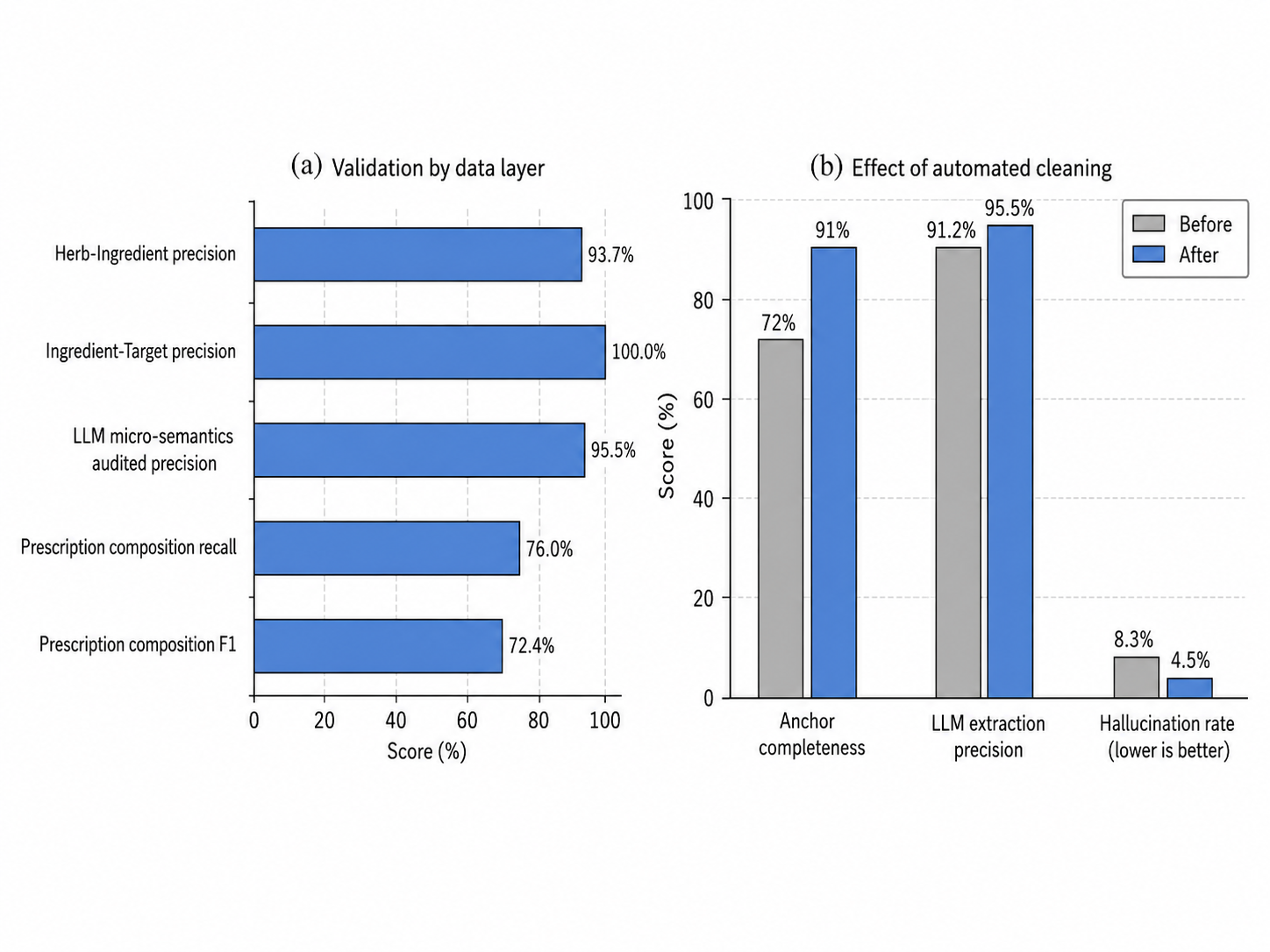}
\caption{Data quality and cleaning effectiveness by validation layer and before/after cleaning metric.}
\label{fig:data-quality}
\end{figure}

\textbf{Pipeline effectiveness.} Batch parallelization reduces cleaning time by about 60\%. After cleaning, anchor-field completeness increases from 0.72 to 0.91, LLM extraction precision improves from 91.2\% to 95.5\%, and hallucination drops from 8.3\% to 4.5\% (Fig.~\ref{fig:data-quality}).

\textbf{Cross-source validation.} Validation against external references yields: herb-ingredient fuzzy precision 93.7\%, ingredient-target precision 100.0\%, and LLM micro-semantics audited precision 95.5\%. Prescription composition recall is 76.0\% (F1 = 72.4\%), reflecting reference formulary incompleteness relative to our multi-source coverage.

These results support two separate claims. First, the graph is clean enough to serve as a retrieval substrate: the most latency-critical molecular and micro-semantic relations pass high-precision validation. Second, recall is not uniformly high because TCM formulary sources disagree in naming, granularity, and composition variants. TCMaster therefore preserves provenance and confidence rather than forcing all sources into a single unqualified truth table.

The cleaning results also clarify the role of LLM extraction. We use LLMs to recover micro-semantic relations that are absent from structured databases, but the graph never treats these triples as first-class authoritative facts without checks. Anchor-field validation removes triples that cannot be linked back to core entities; audit scoring estimates whether the extracted relation is supported by the source text; and the confidence model keeps LLM\_EXTRACTED edges below curated or pharmacopoeia-derived relations. This pipeline turns LLM extraction into a controlled data-ingestion channel rather than an unverified data source.

\subsection{Structural Validation via Link Prediction}

\begin{figure}[htbp]
\centering
\includegraphics[width=\columnwidth]{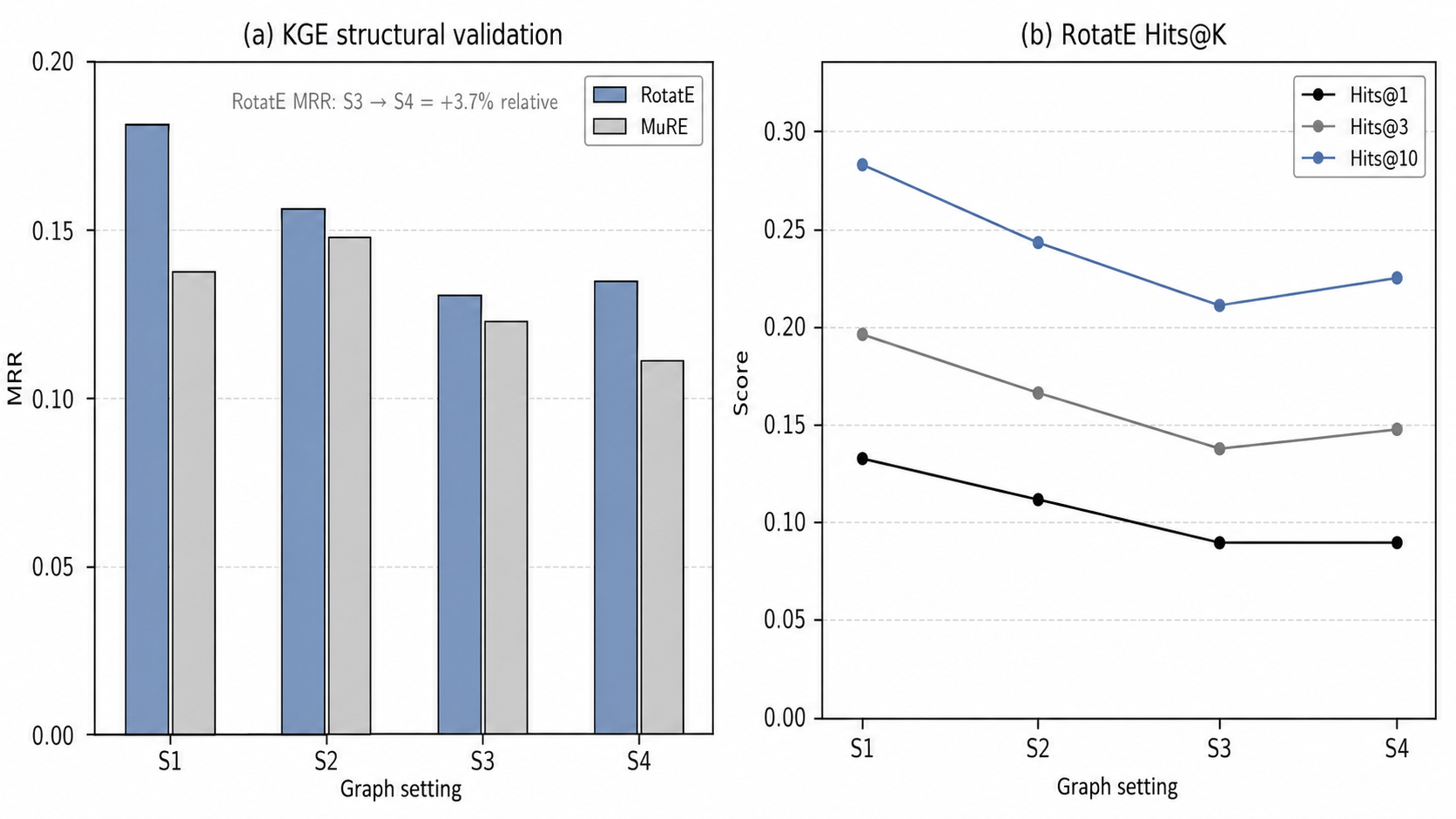}
\caption{KGE structural-validation ablation across S1--S4, with MRR baselines and RotatE Hits@1/3/10.}
\label{fig:kge-ablation}
\end{figure}

We use link prediction as a structural validation probe rather than as a primary system contribution. Figure~\ref{fig:kge-ablation} reports incremental settings S1--S4 using mean reciprocal rank (MRR). S1 (molecular only) MRR = 0.185; S2 (+TCM attributes) MRR = 0.157; S3 (+prescriptions) MRR = 0.131; S4 (+LLM micro-KG) MRR = 0.136.

\textbf{Ablation interpretation.} Adding TCM attributes lowers RotatE MRR from 0.185 to 0.157 because many herbs map to a few controlled attribute nodes, although MuRE improves from 0.142 to 0.149 (+8.0\%), consistent with the layer's tree-like topology. Adding prescriptions further increases candidate-space and relation heterogeneity, reducing RotatE MRR to 0.131. Adding LLM micro-semantics recovers a modest RotatE gain to 0.136 (+3.7\%). We treat KGE as a diagnostic stress test; extraction audit and cross-source validation remain the primary quality evidence.

The main lesson is that structural predictability is not monotonic with graph size. Adding useful domain layers can make the prediction task harder because the candidate space expands and relation semantics diversify. We therefore do not interpret lower MRR as evidence that a layer is harmful. Instead, KGE complements the query benchmarks by showing how each layer changes relational geometry under standard embedding objectives.

\subsection{Downstream Clinical QA (TCMbench)}

We evaluate a downstream KG-RAG application built on TCMaster using TCMbench v3 (2,000 clinical questions across Clinical Prescription, Prescription Logic, and Safety Audit dimensions) and DeepSeek-V3 as the reasoning LLM. TCMbench v3 is our clinical evaluation split built from the submitted artifact's benchmark data, while external TCM QA benchmarks are cited for comparison context~\cite{b30,b33}. Results are shown in Table~\ref{tab:tcmbench}.

\begin{table}[htbp]
\caption{TCMbench accuracy by mode and dimension}
\begin{center}
\begin{tabular}{lrrr}
\toprule
\textbf{Dimension} & \textbf{Baseline} & \textbf{Vector-RAG} & \textbf{KG-RAG}\\
\midrule
Clinical Prescription & 17.9\% & 14.1\% & 45.3\%\\
Prescription Logic & 99.7\% & 99.4\% & 100.0\%\\
Safety Audit & 40.2\% & 44.1\% & 72.2\%\\
\midrule
\textbf{Overall} & \textbf{52.5\%} & \textbf{52.5\%} & \textbf{72.5\%}\\
\bottomrule
\end{tabular}
\label{tab:tcmbench}
\end{center}
\end{table}

KG-RAG improves overall accuracy by 20.0 pp over Baseline and Vector-RAG (72.5\% vs.\ 52.5\%); Wilson intervals are 70.5--74.4\% for KG-RAG and 50.3--54.7\% for the baseline. Clinical Prescription improves by 27.4 pp and Safety Audit by 32.0 pp, while Vector-RAG underperforms Baseline on Clinical Prescription (14.1\% vs.\ 17.9\%) due to context dilution. This is downstream validation of structured KG retrieval, not a new LLM architecture or a direct causal test of the confidence model. We use full-recall retrieval ($\theta=0$); adaptive confidence thresholds remain future work.

The dimension-level results are consistent with the graph structure. Prescription Logic is already near saturated for the baseline, leaving little headroom. Clinical Prescription and Safety Audit require retrieving specific herbs, properties, warnings, and toxicity nodes, where structured paths are more reliable than unstructured passages. The Vector-RAG failure case reinforces the same point: text similarity can retrieve plausible but incomplete passages, while KG retrieval constrains context to typed relations.

We deliberately keep this experiment downstream rather than central. KG-RAG accuracy is not used to tune the optimizer, and the retrieval setting uses full recall with $\theta=0$ so that the result does not conflate confidence filtering with answer generation. Its purpose is to show that the query substrate exposes useful structured context to an application layer. The stronger system claim remains the database claim evaluated in the query benchmarks: confidence and provenance are available at query time, and workload-specific physical structures change latency only when the workload matches their assumptions.

\begin{figure}[htbp]
\centering
\includegraphics[width=\columnwidth]{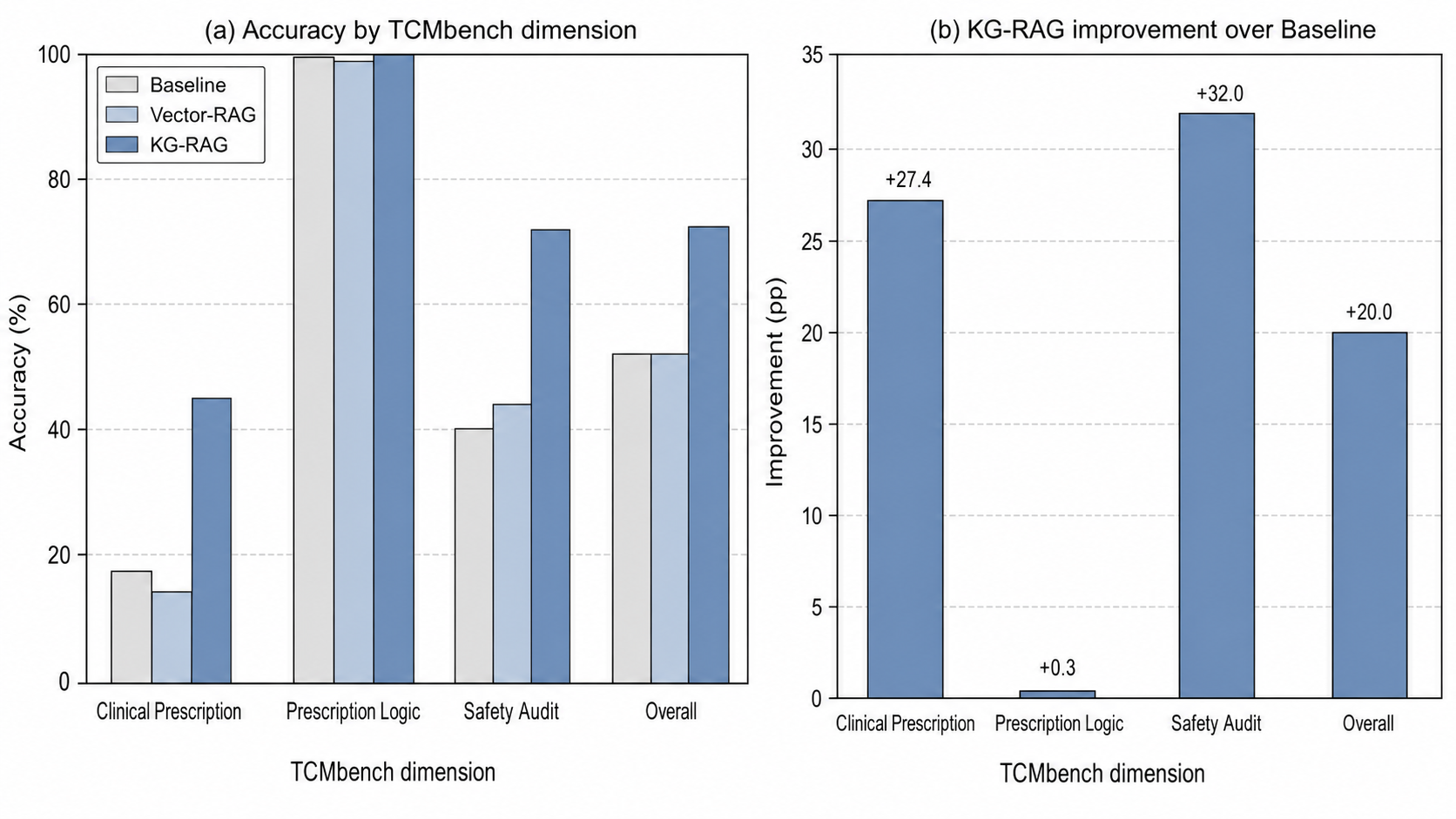}
\caption{Downstream TCMbench validation: accuracy by mode and KG-RAG gains by dimension.}
\label{fig:tcmbench-results}
\end{figure}

\subsection{Confidence Query Analysis}

\textbf{Threshold filtering effect.} EXP-11a evaluates PRODUCT filtering over five workloads and thresholds 0.10--0.45. Filtering is selective: Q1, Q2, Q5, and Q7 retain all results because they traverse mostly AUTHORITATIVE, AUTHORITATIVE\_PHARMA, or CURATED relations, while Q3 mixes curated, molecular, and predicted evidence. At $\theta = 0.18$, Q3 filters 39.3\%; at $\theta = 0.30$, it filters 69.5\%. Thresholds above 0.35 remove all Q3 paths, useful for conservative safety filtering but too strict for exploratory target discovery.

This behavior is desirable for interactive analysis. Raising $\theta$ does not uniformly shrink every query; it primarily affects paths that combine heterogeneous evidence. In practice, a curator can keep authoritative attribute and prescription queries unchanged while tightening mixed molecular paths. The threshold is therefore a workload-specific reliability knob rather than a global quality switch.

\begin{table}[t]
\caption{Filtering ratio (\%) under confidence thresholds}
\label{tab:conf-threshold}
\centering
\footnotesize
\setlength{\tabcolsep}{3pt}
\begin{tabular}{lrrrrr}
\toprule
\textbf{Query} & $\theta=0.10$ & $\theta=0.18$ & $\theta=0.30$ & $\theta=0.35$ & $\theta=0.45$\\
\midrule
Q1 & 0.0 & 0.0 & 0.0 & 0.0 & 0.0\\
Q2 & 0.0 & 0.0 & 0.0 & 0.0 & 0.0\\
Q3 & 0.0 & 39.3 & 69.5 & 100.0 & 100.0\\
Q5 & 0.0 & 0.0 & 0.0 & 0.0 & 0.0\\
Q7 & 0.0 & 0.0 & 0.0 & 0.0 & 0.0\\
\bottomrule
\end{tabular}
\end{table}

\begin{figure}[htbp]
\centering
\includegraphics[width=\columnwidth]{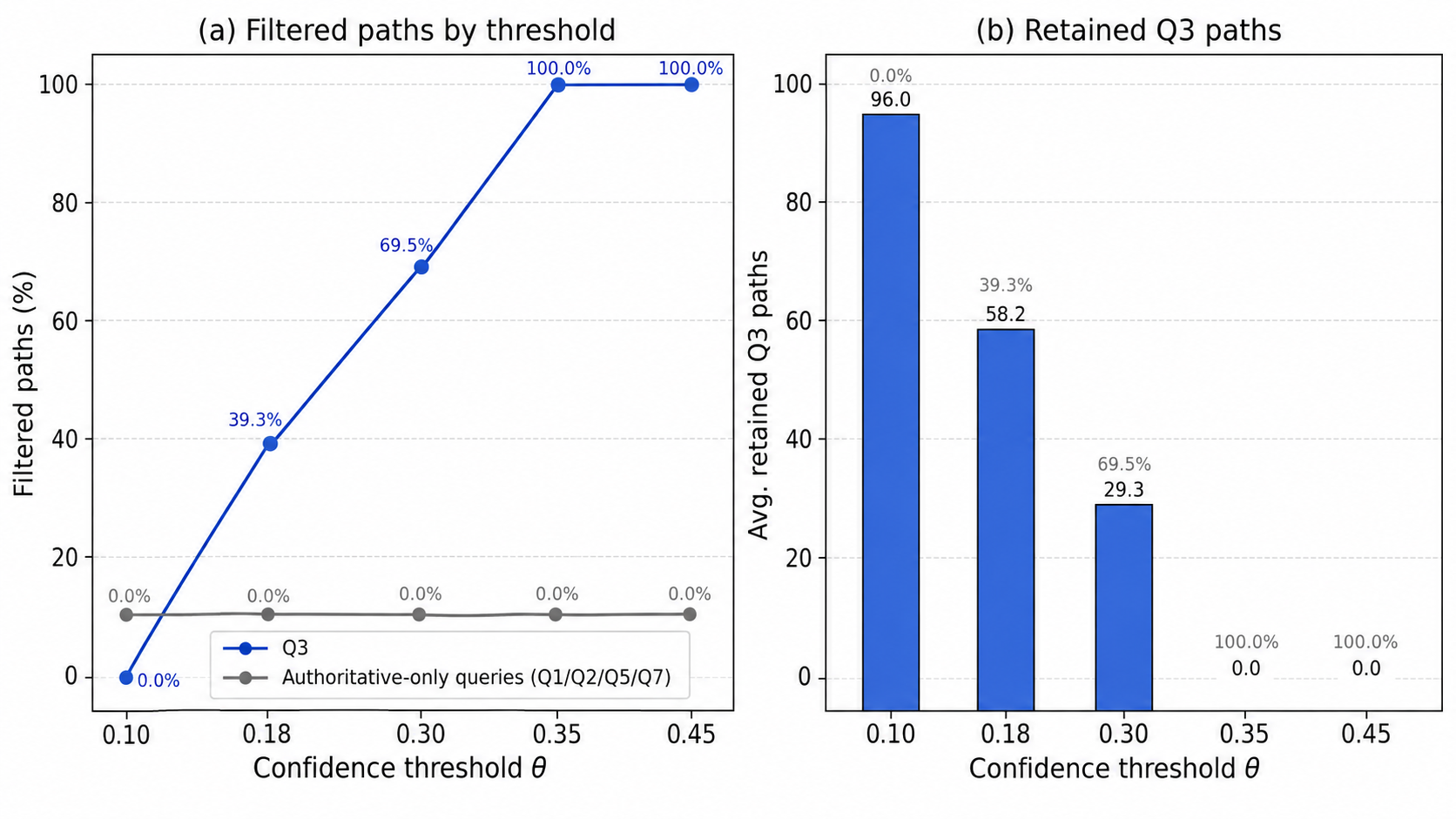}
\caption{Effect of confidence threshold $\theta$ on filtering ratio and recall/precision trade-off.}
\label{fig:confidence-threshold}
\end{figure}

\textbf{Aggregation strategy comparison.} EXP-11b fixes the candidate paths and changes only aggregation. PRODUCT, MIN, and WAVG inspect the same 184.5 average paths but yield mean confidences of 0.0248, 0.1347, and 0.3786, respectively. Thus aggregation is a semantic choice: PRODUCT is conservative, WAVG preserves exploratory recall, and MIN reflects the weakest edge.

The large gap between PRODUCT and WAVG also explains why TCMaster exposes the policy instead of hard-coding one score. Long biomedical paths often combine curated prescription edges with molecular evidence; PRODUCT penalizes such chains strongly, which is appropriate for safety inspection but may suppress useful candidates in discovery tasks. WAVG and MIN let applications choose a less conservative interpretation while keeping provenance visible.

\textbf{Incremental feedback cost.} EXP-11c measures expert confidence updates. A 10-edge batch costs 28.78~ms per edge, while 100-, 500-, and 1,000-edge batches amortize to 5.37, 4.67, and 4.83~ms per edge, supporting small-batch review and larger background curation without full-graph recomputation.

\subsection{Case Study: Prescription Evidence Inspection}

We use \textit{Ephedra Decoction} (\textit{Ma Huang Tang}) to illustrate evidence inspection, not clinical deployment. Given the query ``What are the ingredients, natures/flavors, and toxicity warnings?'', TCMaster traverses Prescription$\rightarrow$Herb, Herb$\rightarrow$Nature/Flavor, and Herb$\rightarrow$Toxicity/Warning paths. With PRODUCT confidence $\geq 0.7$, it returns facts such as Ephedra having nature Hot and flavor Pungent (0.90, AUTHORITATIVE\_PHARMA), an Ephedra toxicity warning (0.85, CURATED), and full path confidence 0.72 for Ephedra Decoction$\rightarrow$Ephedra$\rightarrow$Hot. The query completes in 6.7~ms and supports expert evidence triage without replacing clinical judgment.

This example shows the interaction among the paper's three mechanisms. Confidence filtering controls which paths are shown, provenance labels explain why they are trusted, and physical design keeps the multi-hop inspection interactive. The same workflow can be applied to other prescriptions by changing the start node, while the returned paths remain auditable because they include edge-level source labels.

\subsection{Discussion and Threats to Validity}

TCMaster's gains concentrate where its assumptions hold: confidence heterogeneity, ontology skew, and repeated cross-layer traversal. Shortcuts help Q3/EXP12 because Prescription$\rightarrow$Herb$\rightarrow$Ingredient$\rightarrow$Target paths create large frontiers; direction reversal helps when ontology nodes are selective; PRODUCT filtering affects heterogeneous paths most strongly. Negative results are also informative: native top-$k$ lookup, bitmap predicates, and shortcut-plus-confidence composition do not always beat Neo4j, so TCMaster reports workload-dependent rewrites rather than a universal graph optimizer.

This is the intended system boundary. The contribution is not that every TCM query becomes faster, but that reliability and stable ontology structure can be surfaced to query execution in a controlled way. When a workload lacks those properties, the native graph engine remains competitive and TCMaster should avoid unnecessary rewrites.

Threats to validity are fourfold. Latency is measured on Neo4j 4.4 and one hardware setting; other engines may differ. Query templates reflect our clinical workloads, not all TCM applications. Confidence values are source-category scores rather than learned probabilities. Safety Audit accuracy is insufficient for autonomous clinical use. We mitigate these threats by reporting workload structure, separating techniques, fixing templates and seeds, and releasing scripts and results; the external-validity claim is limited to domains with small controlled ontologies linked to high-cardinality entities.

A further limitation is that the current system uses rule-based rewrite selection. This choice is deliberate for the first artifact because each rule can be inspected and connected to a measured workload. A learned or fully cost-based selector would be a natural next step, but would require a larger training set of query templates and cardinality observations. The current evaluation therefore emphasizes explainable rewrite applicability: each optimization has a stated structural precondition, a measured positive case, and at least one boundary case where it should not be applied.

\section{Related Work}
\label{sec:related}

TCM resources such as TCMSP~\cite{b1}, TCMID~\cite{b2}/TCMID~2.0~\cite{b6}, SymMap~\cite{b3}, HERB~2.0~\cite{b4}, and ETCM~\cite{b5} provide molecular, syndrome/symptom, and prescription data, while surveys note gaps in normalization and reliability annotation~\cite{b19}. Biomedical KGs such as Hetionet and PrimeKG support drug-repurposing integration but do not model TCM attributes, prescriptions, or provenance-aware clinical query processing~\cite{b40,b41}. TCMaster instead targets confidence-annotated path querying and ontology-aware execution over a multi-source TCM KG.

\textbf{Data quality, provenance, and uncertainty.} ActiveClean, HoloClean, and KATARA study cleaning, repair, and validation using model feedback, probabilistic inference, or external knowledge~\cite{b12,b13,b14}. Probabilistic databases, ULDBs, and provenance semirings provide uncertainty and lineage semantics~\cite{b17,b18,b37}. TCMaster takes an operational property-graph point in this space: deterministic cleaning before ingestion, source-level confidence on edges, confidence predicates pushed into Cypher, and path ranking by aggregate confidence for low-latency traversal.

\textbf{Graph query processing and optimization.} Prior work studies graph languages, Cypher, RDF indexing, join ordering, and worst-case optimal joins~\cite{b9,b11,b38,b39,b25,b26,b10}. Beyond labels, predicates, indexes, and join patterns, TCMaster exploits stable domain-structural skew between small ontology attributes and high-cardinality herbs, prescriptions, ingredients, and targets. Direction reversal, compact attribute bitmaps, and materialized cross-layer shortcuts exploit this skew when it matches the workload; native top-$k$ lookup and shortcut-plus-confidence predicates show why rewrites must remain selective.

\textbf{Knowledge graph embeddings and structural validation.} Models such as TransE, RotatE, ComplEx, DistMult, TuckER, MuRE, and R-GCN are widely used for graph representation learning~\cite{b7,b20,b21,b22,b24,b23,b31,b16}. TCMaster uses them only as diagnostic probes: the S1--S4 ablation measures how prescriptions, ontology attributes, and micro-semantics change structural predictability. Since MRR depends on candidate space, relation heterogeneity, and negative sampling, these results complement extraction audits and query benchmarks rather than proving data quality alone.

\textbf{TCM retrieval and KG-RAG.} Retrieval-augmented generation and GraphRAG motivate structured retrieval for knowledge-intensive QA~\cite{b15,b32}; recent TCM work studies LLM-assisted diagnosis, prescription generation, KG-enhanced retrieval, and benchmark construction~\cite{b27,b28,b29,b30,b33}. OpenTCM~\cite{b29} is closest in application motivation, but evaluates an LLM-facing pipeline. TCMaster focuses on the data-management substrate: confidence/provenance edge properties, confidence-bounded traversal, and a separate evaluation of graph-query workloads and downstream KG-RAG accuracy.

\section{Conclusion and Limitations}
\label{sec:concl}

We presented TCMaster, a confidence-aware graph query substrate for multi-source TCM knowledge graphs. By annotating 723K base edges with source-level confidence scores and exploiting ontology skew through direction selection, compact attribute bitmaps, and materialized shortcuts, TCMaster supports millisecond-level multi-hop evidence retrieval. Direction selection improves attribute lookup by a factor of 1.47, materialized shortcuts accelerate high-fanout target counting by a factor of 4.42, and downstream KG-RAG improves clinical QA by 20.0 pp.

\textbf{Limitations.} TCMaster currently uses fixed source-category confidence scores; learning dynamic confidences from user feedback remains future work. The scalability benchmark covers L1--L3, which contain the latency-critical paths used by our workloads, but extending controlled scale-up to L4--L5 would give broader evidence. KG-RAG uses fixed top-3 full-recall retrieval without confidence thresholding, and the case study covers one prescription. Safety-oriented outputs are evidence inspection aids for expert review, not deployable clinical safety decisions. Finally, bitmap filtering and shortcut-plus-confidence predicates do not consistently outperform Neo4j on top-$k$ lookup, motivating future cost-based rewrite selection. These limitations do not change the main claim: confidence and ontology structure can be first-class execution signals for TCM KGs, but rewrite policy depends on workload shape and evidence requirements.

Future work includes dynamic confidence learning, adaptive KG-RAG retrieval, and generalizing ontology-skew optimization to domains with similar hierarchical structure.

\section*{Artifact and Data Availability}

For review, the supplemental artifact is provided as a separate open repository/archive URL with source code, experiment scripts, query templates, configurations, raw/processed results, and a README mapping reported tables to reproduction commands. To avoid requiring reviewers to rerun the 36-hour ETL pipeline, it includes a Neo4j~4.4 dump of the materialized graph, including 28.75M \texttt{REACHES\_TARGET} edges, plus load instructions, validation queries, schema documentation, checksums, and lightweight verification scripts. After publication, the code will be released on GitHub and archived with a persistent snapshot. TCMaster-KG will be available through an open-access browsing website and downloadable Neo4j/CSV/JSONL snapshots with schema documentation and a data card; bulk reuse should use snapshots rather than crawling.

\section*{AI-Generated Content Acknowledgement}

This work uses DeepSeek-V3 for LLM-based micro-semantic extraction and audit scoring as described in Section~\ref{sec:data-integration}. The authors also used Claude (Anthropic) for data processing, experiment scripting, and manuscript editing. All AI-generated content, extracted triples, scripts, and manuscript edits were reviewed and verified by the authors, who remain responsible for the correctness and originality of all content.

\end{document}